\documentclass[10pt,twocolumn,letterpaper]{article}
\usepackage{cvpr}              

\usepackage[dvipsnames]{xcolor}

\usepackage{multirow}
\usepackage{amsmath}

\newcommand*{\affaddr}[1]{#1} 

\newcommand*{\email}[1]{\small{\texttt{#1}}}

\usepackage[pagebackref,breaklinks,colorlinks,]{hyperref}
\def\paperID{0273} 
\def\confName{3DV\xspace}
\def\confYear{2024\xspace}

\title{GAPS: Geometry-Aware, Physics-Based, Self-Supervised \\Neural Garment Draping}
\author{
Ruochen Chen\hspace{1cm}
Liming Chen\hspace{1cm}
Shaifali Parashar
\\
\small{\affaddr{CNRS, Université Claude Bernard Lyon 1, INSA Lyon, École Centrale de Lyon, Université Lumière Lyon 2, LIRIS UMR5205}}\\
\email{\{ruochen.chen, liming.chen, shaifali.parashar\}@liris.cnrs.fr}
}

\begin{document}
\twocolumn[{%
\renewcommand\twocolumn[1][]{#1}%
\maketitle
\begin{center}
    \centering
    \captionsetup{type=figure}
    \includegraphics[width=1\textwidth]{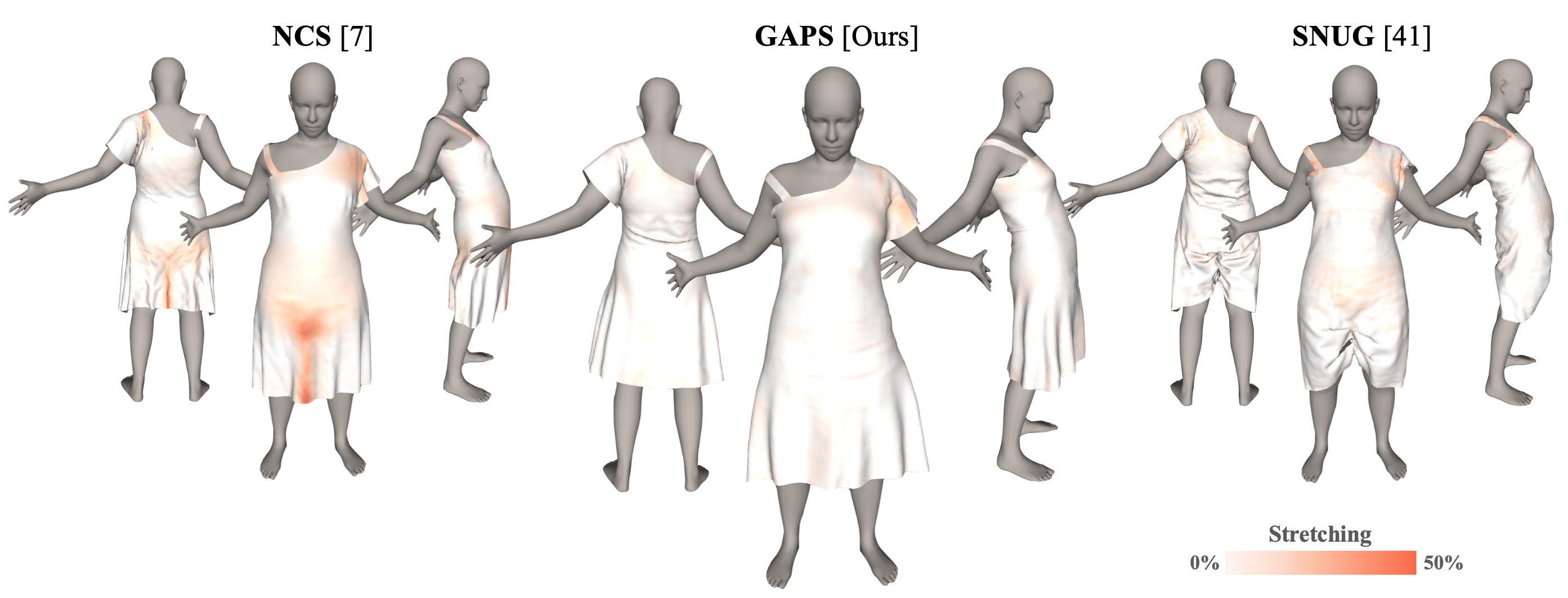}
    
    \captionof{figure}{\textbf{SNUG}~\cite{Santesteban22} and \textbf{NCS}~\cite{Bertiche22} perform self-supervised draping of garments. However, they are prone to unrealistic stretching arising due to their efforts to prevent garment-body collisions. To overcome this issue, we propose \textbf{GAPS} which performs geometrically aware mechanism to drape a garment by stretching only if necessary. 
    This controls body-garment collisions automatically without any post-processing~\cite{Santesteban22} or restrictive measures~\cite{Bertiche22}. Furthermore, our geometry-aware skinning allows better handling of loose garments.} \label{fig:teaser}
\end{center}%
}]

\begin{abstract}
Recent neural, physics-based modeling of garment deformations allows faster and visually aesthetic results as opposed to the existing methods. Material-specific parameters are used by the formulation to control the garment inextensibility. This delivers unrealistic results with physically implausible stretching. Oftentimes, the draped garment is pushed inside the body which is either corrected by an expensive post-processing, thus adding to further inconsistent stretching; or by deploying a separate training regime for each body type, restricting its scalability. Additionally, the flawed skinning process deployed by existing methods produces incorrect results on loose garments.

In this paper, we introduce a geometrical constraint to the existing formulation that is collision-aware and imposes garment inextensibility wherever possible. Thus, we obtain realistic results where draped clothes stretch only while covering bigger body regions. Furthermore, we propose a geometry-aware garment skinning method by defining a body-garment closeness measure which works for all garment types, especially the loose ones. Our code is publicly available at \href{https://github.com/Simonhfls/GAPS}{https://github.com/Simonhfls/GAPS}.
\end{abstract}

\vspace{-10pt}    
\section{Introduction}
\label{sec:intro}
Modeling digital garments for real-time applications, such as video game simulations, fashion design, and e-commerce virtual try-on systems is a challenging yet important problem to solve. Traditionally, clothes are modeled using Physics-Based Simulation (PBS)~\cite{Nealen06,Baraff98,Narain12} of continuum fabric~\cite{Macklin16} or highly detailed yarn~\cite{Cirio14} as per Newton's laws of motion.
While these techniques are capable of delivering precise simulations, their high computational complexity makes them unsuitable for real-time and web-based applications. 

With the success of deep learning in major 3D learning tasks~\cite{Qi17a,Qi17b,Soltani17,Omran18,Richardson16}, supervised learning of garment draping~\cite{Gundogdu20,Tiwari20,Pfaff21,Wang19,Zhang21,Patel20,Corona21} gained interest. They used a neural network fusing a given garment and a body (a joint pose and shape representation generally parametrized using SMPL~\cite{Loper15})  to produce drapings similar to PBS, but at a faster rate. 

A huge downside is that the supervision requires high quality 3D meshes of draped garments over various bodies which not only requires expensive setup (3D scanners and multi-view cameras) but also a huge amount of manual intervention. In addition, all these methods leave significant smoothing artefacts that often result in visually unaesthetic drapings.

Recently,~\cite{Bertiche21} proposed a self-supervised learning of garment draping by enforcing a static physical consistency of garments while training. Inspired by~\cite{Sebastian11}, ~\cite{Santesteban22,Bertiche22} reformulated motion constraints in PBS as an optimization problem and enforced physical consistency on both static and dynamic garment deformations. Not only such a formulation is computationally more efficient than PBS and supervised methods but it also does not incur smoothing artefacts. 
However, a major shortcoming of~\cite{Santesteban22,Bertiche22} is the assumption that the plasticity or elasticity of worn clothes depends solely on material properties. When a garment fits onto a body, it remains inextensible in most parts and stretches only when wrapped over bigger body regions. Current models use only material parameters and limit the strain and bending forces to allow only small changes. This causes two major problems: 1) Unrealistic fittings with abnormal local stretches that are incoherent with the physical nature of clothes, see Fig.~\ref{fig:teaser}. 2) Unsolvable body-garment collisions due to the naive strain/bending minimization. To overcome 2),~\cite{Santesteban22} relies on heavy post-processing that forcefully pushes the collided garment vertex outside the body. This impairs run-time performance and contributes to further incoherent stretching.~\cite{Bertiche22} uses data augmentation to reduce collisions and limits the training to single body shape. This is impractical as it requires to re-train for each body shape. It leads to longer training duration (reported up to 24 hours for loose garments such as dresses).~\cite{Santesteban22} simultaneously trains to drape a garment over several body types. However, it does not handle cloth dynamics well.
In addition, current methods use flawed garment-skinning whose accuracy is dependent on the tightness of the body which degrades their performance
 poorly while draping loose garments as seen Fig.~\ref{fig:teaser}. 

In this paper, we propose \textbf{GAPS}: a geometry-aware, physics-based, self-supervised neural garment draping method to fix the above-mentioned problems. Geometry-based deformation modeling is widely popular in many computer vision problems such as 3D reconstruction~\cite{Parashar20,Parashar21,Bednarik20,Perriollat08} 
 and shape matching~\cite{Bednarik21,Donati20}. To our best knowledge, \textbf{GAPS} is the first neural method that combines physics-based formulation with geometrical constraints on deformations to force collision-aware garment inextensibility. It is built upon~\cite{Santesteban22} and imposes inextensibility by preserving local distances and areas, which is only progressively relaxed in case of body-garment collisions. Consequently, it replicates the true cloth behaviour: maintains inextensibility, stretches only over bigger body regions and does not penetrate the body.
Furthermore, we propose a novel, geometry-aware skinning method which is suitable for all garment types as seen Fig.~\ref{fig:teaser}.
Existing point-based methods \cite{Ma2021, Chen2021} learn a backward transformation field between posed body and canonical pose which fails on loose garments such as skirts. More advanced strategies such as \cite{Ma2022} learn implicit neural garment body interactions but they don't generalise well. Contrary to these works, our proposed skinning is rather simple and able to generalise well.
Our experiments demonstrate a significant qualitative and quantitative improvement over state-of-the-art methods. 

\section{Related Work}
 We classify existing cloth deformation methods into physics-based and learning-based methods.

\noindent \textbf{Physics-based methods.}
Elastic continumm models~\cite{Baraff98} have been widely popular for modeling cloth deformations. However, their computational complexity has been an issue. Many simplifications such as reformulation of elastic model with Finite Element Method~\cite{Kim20}, faster simulations using mass-spring model~\cite{Liu13}, motion constraints~\cite{Macklin16} have been proposed to reduce the computational complexity at a slightly compromised accuracy. To maintain cloth inextensibility,~\cite{Bender08} proposed an impulse-based constraint to minimize changes in lengths. However, this constraint was posed everywhere on the cloth and did not yield a significant improvement.~\cite{Provot95,Bridson02} proposed an iterative refinement as post-processing to minimize changes in edge lengths taking into account the collisions, friction and contacts. Unlike~\cite{Bender08}, we impose inextensibility into the geometrical structure through local covariance matrices, which are widely used to compute surface normals and variations~\cite{Pauly02}. We allow a gradual relaxation on the constraint whenever the draped garment finds itself inside the body; thus imposing it locally.

\noindent \textbf{Learning-based methods.}
Despite several improvements, the computational complexity of PBS methods remains unsuitable for fast or real-time applications.
Learning-based methods, on the other hand, yield a fast inference. Most methods focus on supervised learning. From a large dataset of a garment draped over several bodies,~\cite{Bertiche20,Patel20,Santesteban19,Zhang21,Wang19,Lahner18} learn a parametric garment deformation. Instead of solving for the garment vertex positions at each time step between the garment at rest and the desired pose, they use ground truth supervision to estimate a single function that directly estimates the positioning of draped garment over the input body by taking into account the local and global characteristics of the body and the garment. More advanced versions~\cite{Gundogdu19,Gundogdu20} of these architectures explicitly model the skin-cloth interactions and improve the performance by incorporating local feature pooling. Since these methods are designed to learn a global representation, finer details such as wrinkles are barely produced which leads to plastic-like garment drapings. Most importantly, gathering such a large amount of data is a tedious task: be it from a PBS described above or a multi-camera setup. In addition, training on these large datasets is computationally expensive. 

Recently,~\cite{Bertiche21,Santesteban22,Bertiche22} proposed self-supervised learning of garment deformations which do not suffer from the drawbacks of the above-mentioned methods. While~\cite{Bertiche21} can model only static deformations,~\cite{Santesteban22,Bertiche22} can model both static and dynamic cloth deformations by imposing PBS as an optimization problem. Consequently, the error margin is much higher on these methods and only material-based control over strain and bending forces produces unrealistic stretches. 
~\cite{Bertiche21} proposed a cloth consistency loss to minimize edge differences and maintain smoothness by minimizing Laplacians in local neighbourhood.~\cite{Bertiche22} also used this cloth consistency model but it is uniformly applied everywhere on the garment even in the areas where the garment needs to stretch in order to fit. Instead, we control garment inextensibility realistically by stretching progressively, only when necessary.
As a result, we do not require to impose any schemes to avoid body-garment collisions. 
While~\cite{Santesteban22} fails on most loose-fitted garments such as dresses due to its flawed skinning, our proposed geometry-aware skinning allows to handle all types of garments easily. We will discuss it further in the methodology section.

\section{Geometry-Aware Modeling}

We now discuss our main contributions. First, we describe geometry-based deformation modeling which together with the existing physics-based modeling allows us to learn true cloth behaviour. Then, we describe the geometry-aware method to compute body-participation in garment dynamics to compute garment skinning. This allows efficient draping of all types of garments.

\subsection{Garment Deformation Modeling}
\label{sec:inext_model}
Even if the garments are made of highly stretchable material, they stretch only while fitting over larger body regions. Depending on the body shape and the dynamic motion, the stretching is generally localized to small areas while the majority of the draped garment remains inextensible; thus preserving local lengths, angles and areas. Geometrically, surfaces that deform while preserving these local properties are defined using \emph{isometry}~\cite{Lee03}: a geodesic-preserving map on smooth manifolds. It is imposed by preserving \emph{metric tensors} which are locally defined using surface jacobians. While this differential modeling delivers promising results, the computation of local jacobians and hessians leads to slow neural networks~\cite{Bednarik20,Bednarik21,Das22}, which is not desirable. We represent garments using meshes which are a discrete representation of a continuous space. Therefore, we relax isometry by enforcing inextensibility locally on each mesh vertex. Given  
$\mathcal{X}=(x_i)_{i=1}^N$, 
a mesh containing $N$ vertices $x_i$ that transforms into 
$\mathcal{Y}=(y_i)_{i=1}^N$,
a naive inextensibility imposition would be force each edge around $y_i$ within its local neigborhood $y_j, j\in \mathcal{N}_i $ to comply with the  corresponding one  around $x_i$ in terms of length. So, we force that 
\begin{equation}
    D_i = \sum_{j \in \mathcal{N}_i} (\|x_i-x_j\|- \|y_i-y_j\|)^2 =0.
\end{equation}
While imposing this constraint minimizes inextensibility, it delivers jittery results. This is because under this scheme, each edge length will be optimized independently causing slight jitters. 

Instead, we impose a restriction on covariance matrix. For $x_i$, it is given by

\begin{equation}
\mathbf{C}_{x_i} = \frac{1}{|\mathcal{N}_i|} \sum_{j\in \mathcal{N}_i} (x_j - \overline{x}_i)(x_j - \overline{x}_i)^\top,
\end{equation}
where $\overline{x}_i$ is the mean of $x_j$. We  impose inextensibility as local rigidity around each vertex. Given that $y_i = \mathbf{R}x_i + \mathbf{T}$, where $\mathbf{R}$ and $\mathbf{T}$ are a 3D rotation and a translation respectively, we assume that $y_j = \mathbf{R}x_j + \mathbf{T}$.
Thus, we obtain
\begin{equation}
\label{eq:covariance_y}
\mathbf{C}_{y_i} = 
 \frac{1}{|\mathcal{N}_i|} \sum_{j\in \mathcal{N}_i} (\mathbf{R}x_j - \mathbf{R}\overline{x}_i)(\mathbf{R}x_j - \mathbf{R}\overline{x}_i)^\top
= \mathbf{R}\mathbf{C}_{x_i}\mathbf{R}^\top.
\end{equation}

Using SVD, we can write $\mathbf{C}_{x_i}= \mathbf{U}\mathbf{S}_{x_i}\mathbf{U}^\top$. It can be easily verified that $\mathbf{C}_{y_i}= \mathbf{R}\mathbf{U}\mathbf{S}_{x_i}\mathbf{U}^\top\mathbf{R}^\top = \mathbf{V}\mathbf{S}_{x_i}\mathbf{V}^\top$. Thus inextensibility implies preservation of singular values of corresponding covariance matrix. 
 It implies that $\sigma_i \in \mathbf{S}_{x_i}=diag(\sigma_1,\sigma_2,\sigma_3)$ will satisfy the characteristic polynomial of $\mathbf{C}_{y_i}$, i.e, 
 
\begin{equation}
    \label{eq:local_inext}
    \det(\mathbf{C}_{y_i} - \alpha \sigma_i \mathbf{I}_{3\times3})=0,
\end{equation}

where $\mathbf{I}_{3\times3}$ is identity and $\alpha =1$. We use the above constraint to enforce realistic cloth behaviour. It has 3 benefits: 1) $\sigma_i$ are derived from template garment and therefore, can be precomputed.
2) We avoid computing singular values of the garment in current timestep during optimization, which is expensive and may suffer from numerical instabilities due to vanishing gradients.
3) We can control the degree of stretching just by changing $\alpha$.
In section~\ref{sec:method}, we will explain how $\alpha$ can be progressively changed to stretch the garment in order to avoid body collisions.

\begin{figure}[htbp]
  \centering
   \includegraphics[width=1\linewidth]{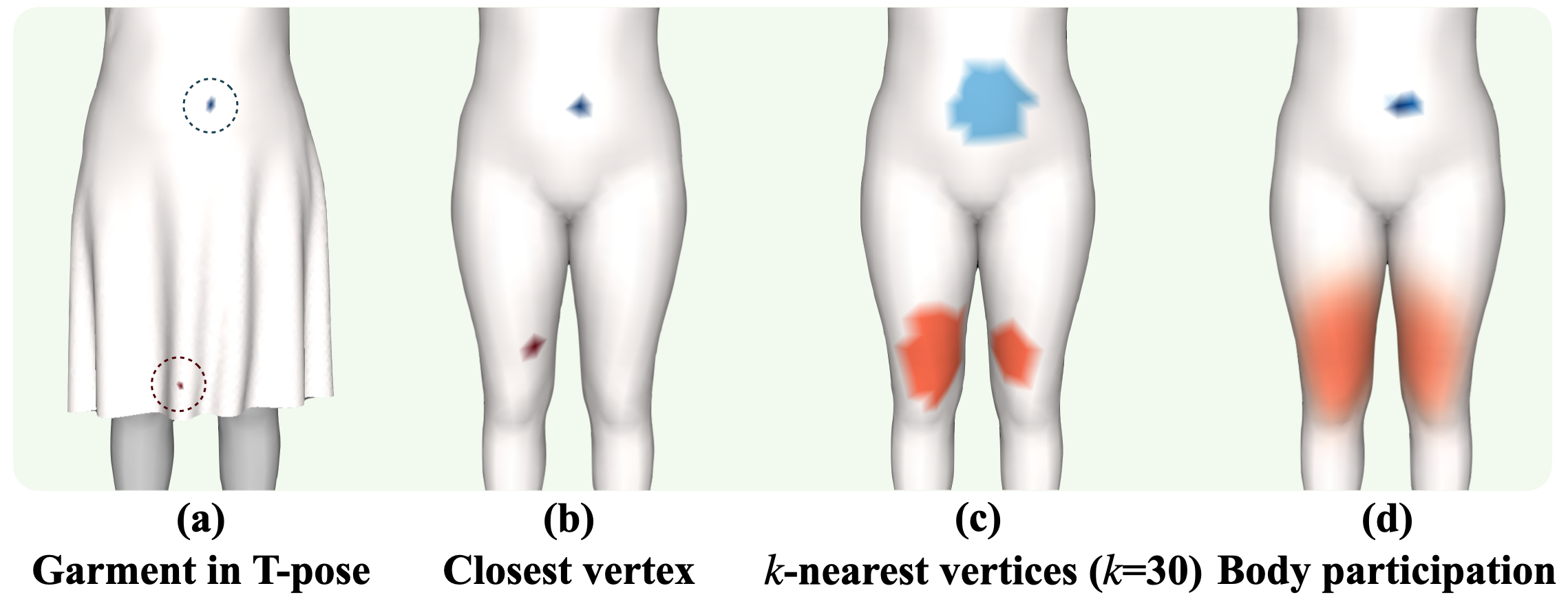}

   \caption{Body-participation computation for garment vertices close (in blue) or far (in red) to the body.}
   \label{fig:blend_weight_contribution}
   \vspace{-10pt}
\end{figure}

\subsection{Body-Participation In Garment Dynamics}
\label{sec:blend weight}

In order to dynamically fit a garment over a body, it is required to compute entire body's participation towards the movement of each garment vertex. As described in~\cite{Loper15}: for each body vertex, it can be computed as a weighted blend of body-joints, with weights describing the impact of body-joints towards the motion at a given vertex. Assuming that the garments follow body motions,~\cite{Bertiche21, Santesteban22, Bertiche22} assign each garment vertex the blending weights of its nearest body vertex directly. 
While this may be effective for tightly-fit garments such as T-shirts, it is extremely suboptimal for looser garments like dresses.
In order to compensate,~\cite{Bertiche22} applies iterative Laplacian smoothing of the weights of each garment vertex w.r.t. its neighbours. On a loose garment, there is a high possibility for a garment vertex and its neighbours to match to the same body vertex, causing undesired artefacts.

To overcome these drawbacks, we introduce a simple, geometry-aware computation of blend weights premised on two fundamental observations: 1) For a given garment vertex, the influence of the blend weights of a body vertex is negatively correlated with their respective Euclidean distances. 2) On loose garments, the blend weights are less likely to depend on the proximate body vertex. We use the Gaussian Radial Basis Function (RBF) to compute the participation of each body vertex to a specific garment vertex. We write a Gaussian RBF kernel as
\begin{equation}
\varphi(r) = e^{-\frac{r^2}{k m_i^2}},
\label{eq:rbf}
\end{equation}
where  $m_i$  is the Euclidean distance between a garment vertex $g_i \in \mathcal{G}$  and its nearest body vertex. Then, 
we define a participation matrix $\mathcal{P}$  where each element  $p_{ij}$  signifies the degree of participation of  body vertex $b_j \in \mathcal{B}$ in $g_i$ as
\vspace{-3pt}
\begin{equation}
    \mathcal{P} = \begin{bmatrix}
    \varphi(d_{11}-m_{1}) & \dots  & \varphi(d_{1N_b}-m_{1})  \\
    \vdots  & \ddots & \vdots \\
    \varphi(d_{N_g1}-m_{N_g})  & \dots  & \varphi(d_{N_gN_b}-m_{ N_g}) 
  \end{bmatrix}
\end{equation}

where $d_{ij}$ represents the Euclidean distance between $g_i$ and $b_j$, $N_g$ and $N_b$ are the number of elements in $\mathcal{G}$ and $\mathcal{B}$ respectively and $k=0.5$.
The new garment blend weights, $\tilde{\mathcal{W}}$, are then given by 
\begin{equation}
\tilde{\mathcal{W}} = \text{normalize}(\mathcal{P} *\mathcal{W}),
\end{equation}
where $\mathcal{W}$ are the blend weights described in~\cite{Loper15}. The normalization makes the cumulative sum of each row to 1.

~\cref{fig:blend_weight_contribution} shows the body participation measures for garment vertices that drape close (in blue) or far (in red) to the body. Picking the closest body vertex and considering its blend weights might work for garment vertices close to the body. However, for the vertices on looser regions, it is clear that motion dynamics is dependent on a larger body region. Naively choosing k-nearest neighbours might work for vertices on looser regions but not on the tight ones. Our simple and effective body-participation method can automatically find the best-fitting region for any type of garment, irrespective of its topology which allows flawless garment skinning.

\section{Method}\label{sec:method}
We integrate our geometric formulations with PBS principles to improve the performance of learning-based cloth simulations. 
\subsection{Model}
\begin{figure}[htbp]
  \centering
\includegraphics[width=1\linewidth]{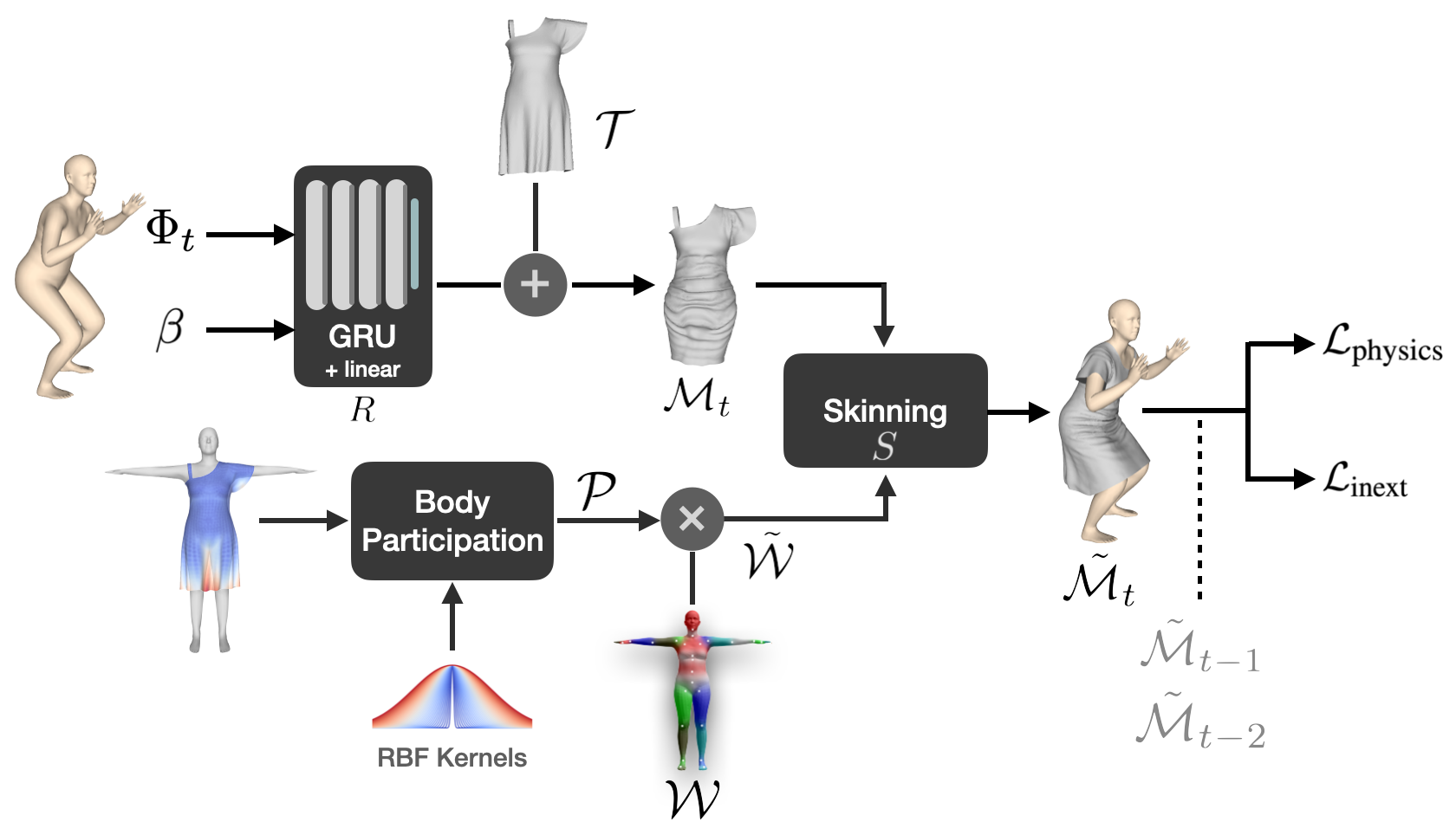}
   \caption{Overview of our method. We added RBF based Body Participation module on~\cite{Santesteban22} to estimate blend weights based on relative garment-body geometry. }
   \label{fig:architecture}
\end{figure}

Our model is developed within an unsupervised learning framework proposed in~\cite{Santesteban22}. It incorporates a Gated Recurrent Unit (GRU) module $R$ outputting vertex displacement relative to the template garment $\mathcal{T}$ (unposed and undeformed). This GRU module consists of four layers, each yielding an output size of 256, with $\tanh$ activation functions. After combining it with $\mathcal{T}$ to obtain the unposed, deformed garment ($\mathcal{M}_t$), we subsequently apply the skinning function $S$ which articulates $\mathcal{M}_t$. The garment blend weights $\tilde{\mathcal{W}}$ for skinning are estimated using Body-Participation module, described in section~\ref{sec:blend weight}.
The network is trained by optimizing a composite loss function, which includes our newly proposed inextensibility loss (described in section~\ref{sec:inext_model}) in addition to the PBS losses. The architecture of our model is depicted in~\cref{fig:architecture}.

The garment is modeled as
\begin{equation}
\begin{aligned}
  &\mathcal{M}_t =\mathcal{T} + R(\Phi_t, \beta), \\
  &\tilde {\mathcal{M}_t} = S(\mathcal{M}_t, J(\beta), \theta_t, \tilde{\mathcal{W}}),
  \end{aligned}
\end{equation}
where the regressor $R$ takes as input the body shape $\beta$ and the motion parameter $\Phi_t=\{\theta_t, v_t\}$ containing the current body pose $\theta_t$ and global velocity $v_t$ of the root joint. The skinning function $S$ articulates $\mathcal{M}_t$ based on joint locations $J(\beta)$, current body pose and the estimated garment blend weights $\tilde{\mathcal{W}}$ using the Body-Participation module.

In order to learn the garment deformations, we know that the physical equilibrium of forces related to strain, gravity, collision and bending (defined in PBS) should be achieved. Like~\cite{Santesteban22}, we enforce them as losses to be minimized. We express them as

\textit{1) Strain.} We use the Saint-Venant–Kirchhoff (StVK) material model for simulating deformation. It is given by
  \vspace{-5pt}
\begin{equation}
\label{eq:strain}
  \mathcal{L}_{\text{strain}} = \mu \| \epsilon_{\text{stvk}} \|_{\text{Fro}}^2 + \frac{\lambda}{2} \mathrm{tr}(\epsilon_{\text{stvk}})^2,
\end{equation}

where $\epsilon_{\text{stvk}} = \frac{1}{2} (\mathbf{F}^\top\mathbf{F} - \mathbf{I})$ represents the Green-Lagrangian strain tensor, where $\mathbf{F}$ is the deformation gradient, and $\mathbf{I}$ is the identity matrix.  $\mu$ and $\lambda$ are the Lamé coefficients reflecting the material properties. 

\textit{2) Gravity.}
We incorporate gravity by minimizing the potential energy of the garment, given by
\begin{equation}
  \mathcal{L}_{\text{gravity}} = \sum_{\text{vertices}}-m g^{\top} x,
\end{equation}
where $m$ is the particle mass and $g$ is the gravitational acceleration.

\textit{3) Collision.} It penalizes penetration between the body and the garment. For each garment vertex, it is given by
\begin{equation}
   \mathcal{L}_{\text{collision}} =  \sum_{\text{vertices}} k_c * d_{c}^3,
\end{equation}
where $k_c$ is a balancing factor, and $d_{c}=\max(\epsilon - d(x), 0)$ quantifies the degree of interpenetration. $d(x)$ is the signed distance between garment vertex and body surface, 
 and $\epsilon$ is a small positive constant introduced to enhance stability.

\textit{4) Bending.} We use a different formulation than~\cite{Santesteban22,Bertiche22}. We express it as a balance between smoothness in local garment areas and global coherence with the original template (to retain natural drape characteristics). It is given by
\begin{equation}
  \mathcal{L}_{\text{bending}} = \sum_{\text{edges}} k_{b} \frac{l^2}{8a} ( \alpha(\theta_t - \theta_r)^2 + (1-\alpha)\theta_t^2),
\end{equation}
where $k_b$ is the bending stiffness, $\theta_{t}$ denotes the dihedral angle of the deformed garment, $\theta_r$ signifies the corresponding dihedral angle in the rest state of the template garment, $l$ is the common edge length and $a$ is the summation of the areas of both faces. The balancing coefficient $\alpha \in [0, 1]$ is proportional to the distance between the garment and the current edge in the rest pose.

\textit{5) Inertia.} In order to convert PBS to an optimisation problem,~\cite{Santesteban22} proposed the intertia loss. It is given by
\begin{equation}
 \mathcal{L}_{\text{inertia}} = \sum_{\text{vertices}} \frac{1}{2\Delta t^2} m (x-x^{[t-1]}- \Delta t v^{[t-1]})^2,
\end{equation}
where $\Delta t$ is the simulation time step, $x^{[t]}$ and $x^{[t-1]}$ specify the particle's position at times $t$ and $t-1$, respectively. 
As suggested in \cite{Bertiche22},
we do not to back-propagate $\mathcal{L}_{\text{inertia}}$ through previous frames.

Together, all these physical losses are given by
\begin{equation}
  \mathcal{L}_{\text{physics}} = \mathcal{L}_{\text{strain}}+ \mathcal{L}_{\text{bending}}+ \mathcal{L}_{\text{collision}}+ \mathcal{L}_{\text{gravity}}+
  \mathcal{L}_{\text{inertia}}.
\end{equation}

\noindent \textbf{Inextensibility Loss.} Our novel geometric loss that enforces inextensibility (as per eq.~\ref{eq:local_inext}) within the one-ring local area associated with each vertex, is given by
\vspace{-5pt}
\begin{equation}
\label{eq:inext_loss}
 \mathcal{L}_{\text{inext}} = k_i  \sum_{\text{vertices}} \sum_{j=1}^3 |\det(\mathbf{C}-k_{\text{ext}}\sigma_{j}\mathbf{I}_{3\times3})|,
\end{equation}
where $k_i$ is a balancing factor, $\mathbf{C}$ is the covariance matrix within 1-ring neighbourhood and $\sigma_j$ are the singular values of covariance matrix of the 1-ring neighbourhood around the corresponding point in garment template $\mathcal{T}$. 
$k_{\text{ext}}$ decides the degree of extension: $k_{\text{ext}}=1$ indicates inextensibility and $k_{\text{ext}}>1$ indicates stretching. Given that garment stretches only while covering large body regions, $k_{\text{ext}}$ should be 1 and changed only if the body-garment collision is detected.
Moreover, to not be affected by slight perturbations, the garment should only be gradually stretched
upon collision detection. Considering this, we write
\vspace{-5pt}
\begin{equation}
k_{\text{ext}} = 1 + \min(10d_{c}, 0.03) \times \min(e, 100),
\end{equation}
where $e$ is the current epoch. We first allow the network to stabilize through 100 epochs and then gradually stretch the garment locally in case of body-garment collisions.
    
The overall loss is
\begin{equation}
  \mathcal{L} =  \mathcal{L}_{\text{physics}} + \mathcal{L}_{\text{inext}}.
\end{equation}
\begin{figure}[htbp]
  \centering
\includegraphics[width=1\linewidth]{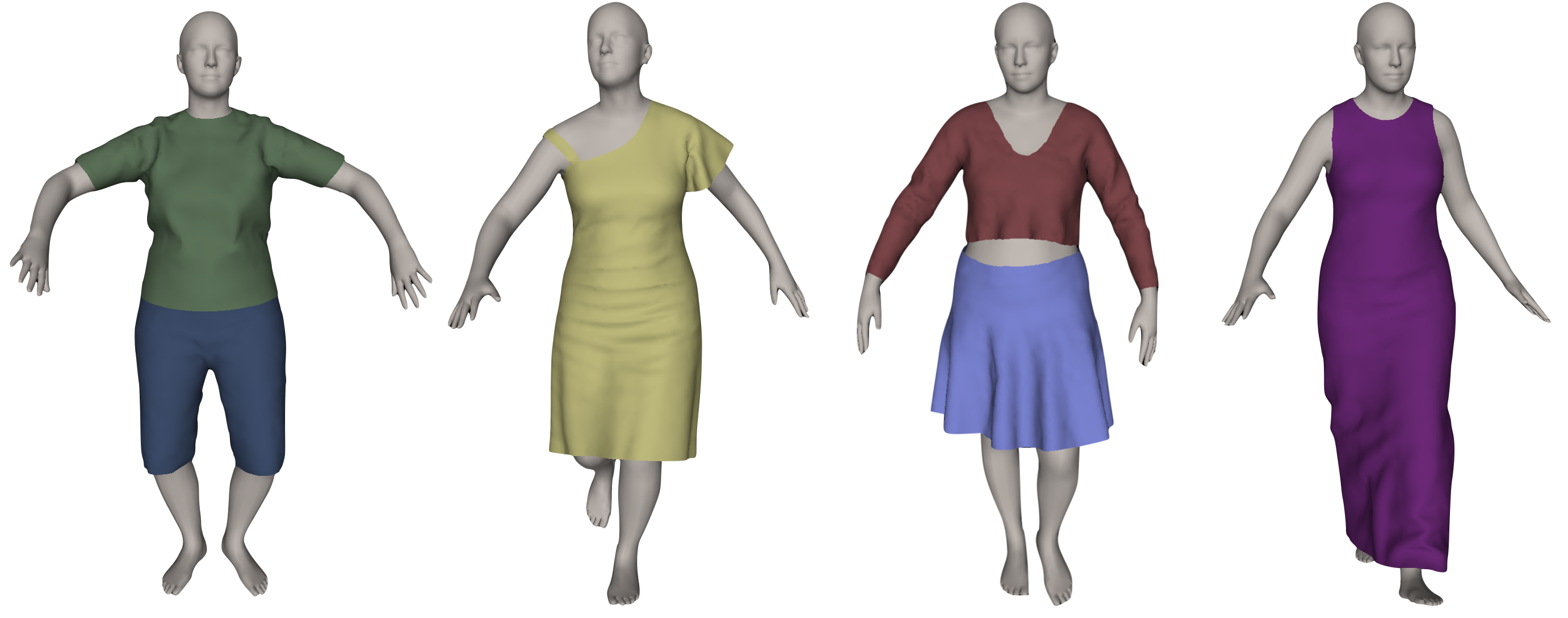}
   \caption{T-shirt, Dress, Skirt, and Gown draped using \textbf{GAPS}. }
   \label{fig:four_garments}
   \vspace{-10pt}
\end{figure}

\vspace{-15pt}
\section{Experiments}

\noindent \textbf{Training.}
During the training phase, the Gated Recurrent Unit (GRU) layers are initialized with random hidden states. The batch size is 128, the learning rate is set to $1e{-3}$ for the initial 10 epochs and $1e{-4}$ for the rest. The balancing weight for inextensibility loss $k_i$ is set to $2e8$. All other hyperparameters as well as material parameters are set according to~\cite{Santesteban22}. We randomly sample the shape parameters from  $\mathcal{U}(-2, 2)$ for each batch. We select 60 train sequences and 5 validation sequences from AMASS dataset~\cite{Mahmood19} containing around 11000 poses. We compare with
\textbf{SNUG}~\cite{Santesteban22} and \textbf{NCS}~\cite{Bertiche22}: the only state-of-the-art methods closest to ours.
For a fair comparison, we train them on the same motion sequences as ours. \textbf{GAPS} and \textbf{SNUG} are trained on variable bodies whereas \textbf{NCS} is trained on a single body. The training code for \textbf{SNUG} is unavailable so we have implemented it. Our training losses are slightly better than the reported numbers in the paper.

\noindent \textbf{Error metrics.} We report $\varepsilon_{\text{e}}$ and $\varepsilon_{\text{a}}$ as the mean difference (in $\%$) between the edge lengths and areas between the template and the draped garment.
We report $\varepsilon_\text{c}$ as the  $\%$ of draped garment vertices colliding with the body.

\begin{figure*}[t]
  \centering
\includegraphics[width=\linewidth]{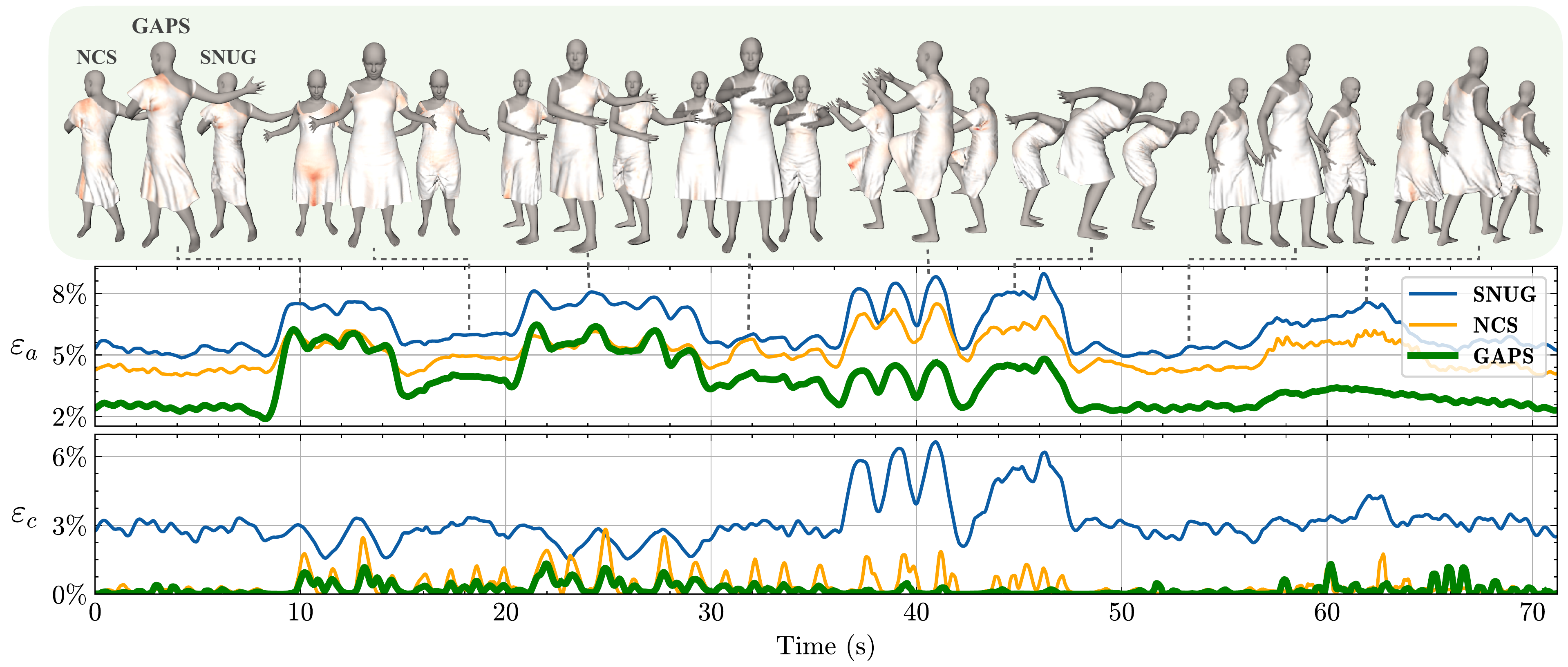}
   \caption{Results on AMASS sequence 86\_07 with Dress. \textbf{GAPS} shows the best performance with minimal collision and realistic stretching (indicated in orange). }
      \label{fig:pct_area} 
      \vspace{-20pt}
\end{figure*}

\noindent \textbf{Quantitative Evaluation.}
We consider the validation sequence AMASS 86\_07 on a single body and drape four garments shown in~\cref{fig:four_garments}. In general, each garment should be able to fit on this body without stretching. However, due to the dynamic motion of the body, stretching is inevitable.~\cref{tab:isometry_1} shows the results. \textbf{GAPS} shows the best performance on all datasets. 
The drapings are obtained with minimal stretch and collisions, thanks to our collision-aware imposition of inextensibility and geometry-aware skinning. We remind that \textbf{NCS} performs body-specific draping. It is trained only on the body that we used in this experiment while  others are trained on variable bodies. It performs well  (close to ours) on T-shirt which is a simple, tight-fitting garment.
On other garments, which are all more challenging than T-shirt, it degrades significantly. \textbf{SNUG} performs a post-processing to fix body-garment collisions which makes $\varepsilon_c =0$ but it considerably increases $\varepsilon_a$ and $\varepsilon_e$. We compute all metrics before post-processing, thus evaluating only the method's performance.
 \cref{fig:pct_area} shows the $\varepsilon_a$ and $\varepsilon_c$ on the entire AMASS 86\_07 sequence with a Dress draped on it. 
 Unlike \textbf{NCS}, the stretches produced by \textbf{GAPS} seem to be quite realistic with minimal stretching in loose garment areas.

\begin{table*}
    \centering
    
    \resizebox{\textwidth}{!}{%
    \begin{tabular}{@{}lcccccccccccc@{}}
    \toprule
    & \multicolumn{3}{c}{T-shirt} & \multicolumn{3}{c}{Dress} & \multicolumn{3}{c}{Skirt} & \multicolumn{3}{c}{Gown} \\
    & $\varepsilon_{e}$ & $\varepsilon_{a}$ & $\varepsilon_{c}$ & $\varepsilon_{e}$ & $\varepsilon_{a}$ & $\varepsilon_{c}$ & $\varepsilon_{e}$ & $\varepsilon_{a}$ & $\varepsilon_{c}$ & $\varepsilon_{e}$ & $\varepsilon_{a}$ & $\varepsilon_{c}$ \\
    \midrule
    \textbf{SNUG} & 7.815 & 13.796 & 5.631 (±2.453) & 4.833 & 6.084 & 3.113 (±1.496) & 3.062 & 4.063 & 0.365 (±0.788) & 5.906 & 7.434 & 3.515 (±3.376) \\
    \textbf{NCS}  & 3.615 & 5.311 & 0.522 (±0.513) & 4.164 & 5.391 & 0.368 (±0.556)  & 4.386 & 5.677 & 2.076 (±0.668) & 4.782 & 6.686 & 0.633 (±1.041) \\
    \textbf{GAPS}  & \textbf{3.343} & \textbf{5.290} & \textbf{0.116 (±0.420)} & \textbf{3.051} & \textbf{3.825} & \textbf{0.196 (±0.308)}  & \textbf{2.337} & \textbf{3.019} & \textbf{0.026 (±0.131)} & \textbf{3.639} & \textbf{4.249} & \textbf{0.123 (±0.225)}\\
    \bottomrule
\end{tabular}
}
\caption{Inextensibility and Collision metrics on AMASS 86\_07. \textbf{GAPS} fits all the garments with minimal collisions and minimal garment extensions. 
\vspace{-5pt}
}
\label{tab:isometry_1} 
\end{table*}

\begin{figure}[t]
  \centering
   \includegraphics[width=1\linewidth]{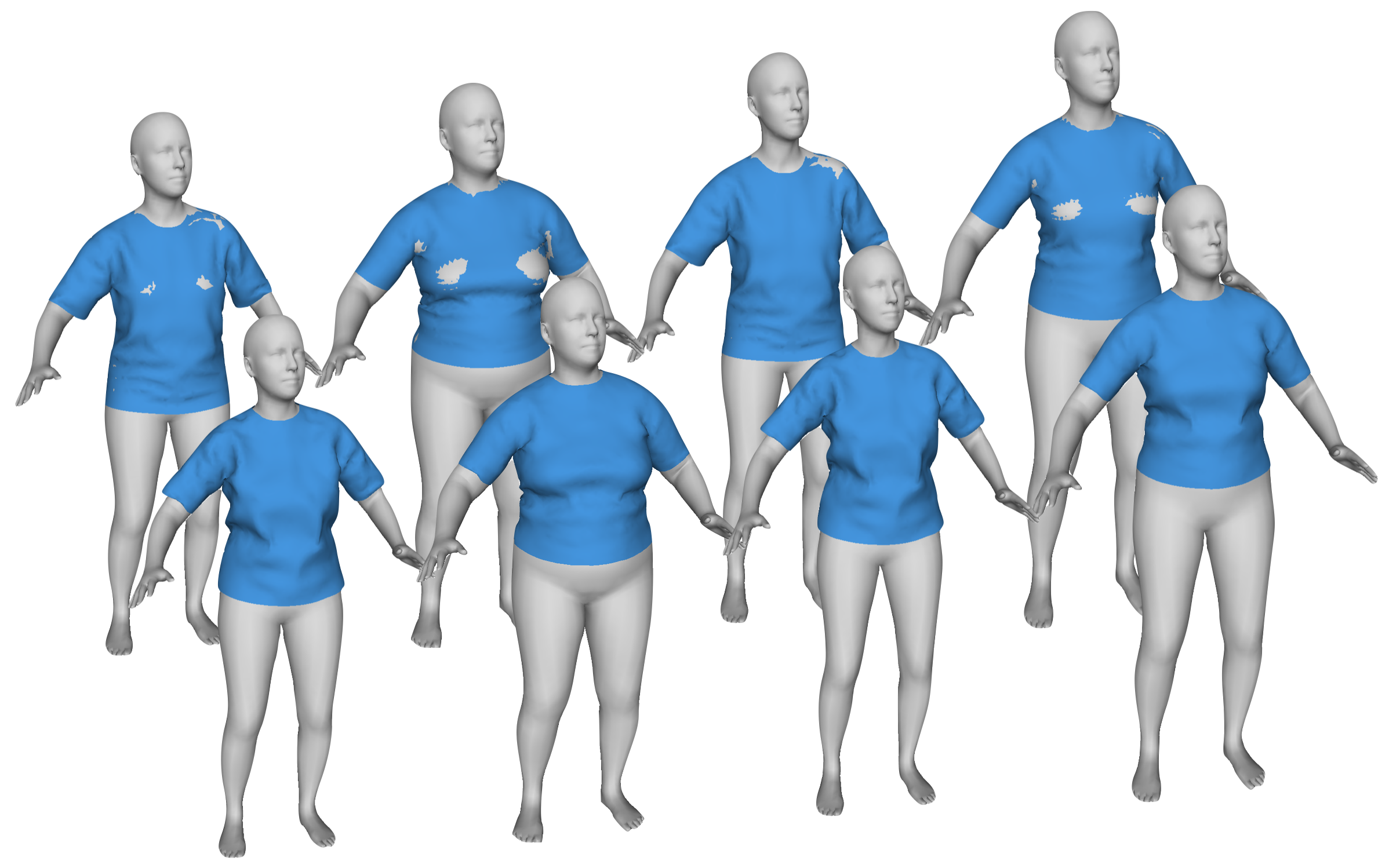}

   \caption{Results on diverse bodies. \textbf{SNUG} (top) shows significant garment penetration while \textbf{GAPS} (bottom) shows remarkable performance. }
   \label{fig:quali_collision}
   \vspace{-10pt}
\end{figure}

\noindent \textbf{Performance on Diverse Bodies.}
Both \textbf{GAPS} and \textbf{SNUG} are trained on variable bodies. We compare their scalability. \cref{fig:quali_collision} shows diverse body shapes where \textbf{SNUG} shows strong body-garment penetrations. It heavily depends on post-processing to generate visually comparable results. In contrast, \textbf{GAPS}' collision-aware inextensibility loss allows visually remarkable results. We have omitted results for extremely thin bodies as they are similar for all methods.

\begin{figure}[htbp]
  \centering
   \includegraphics[width=1\linewidth]{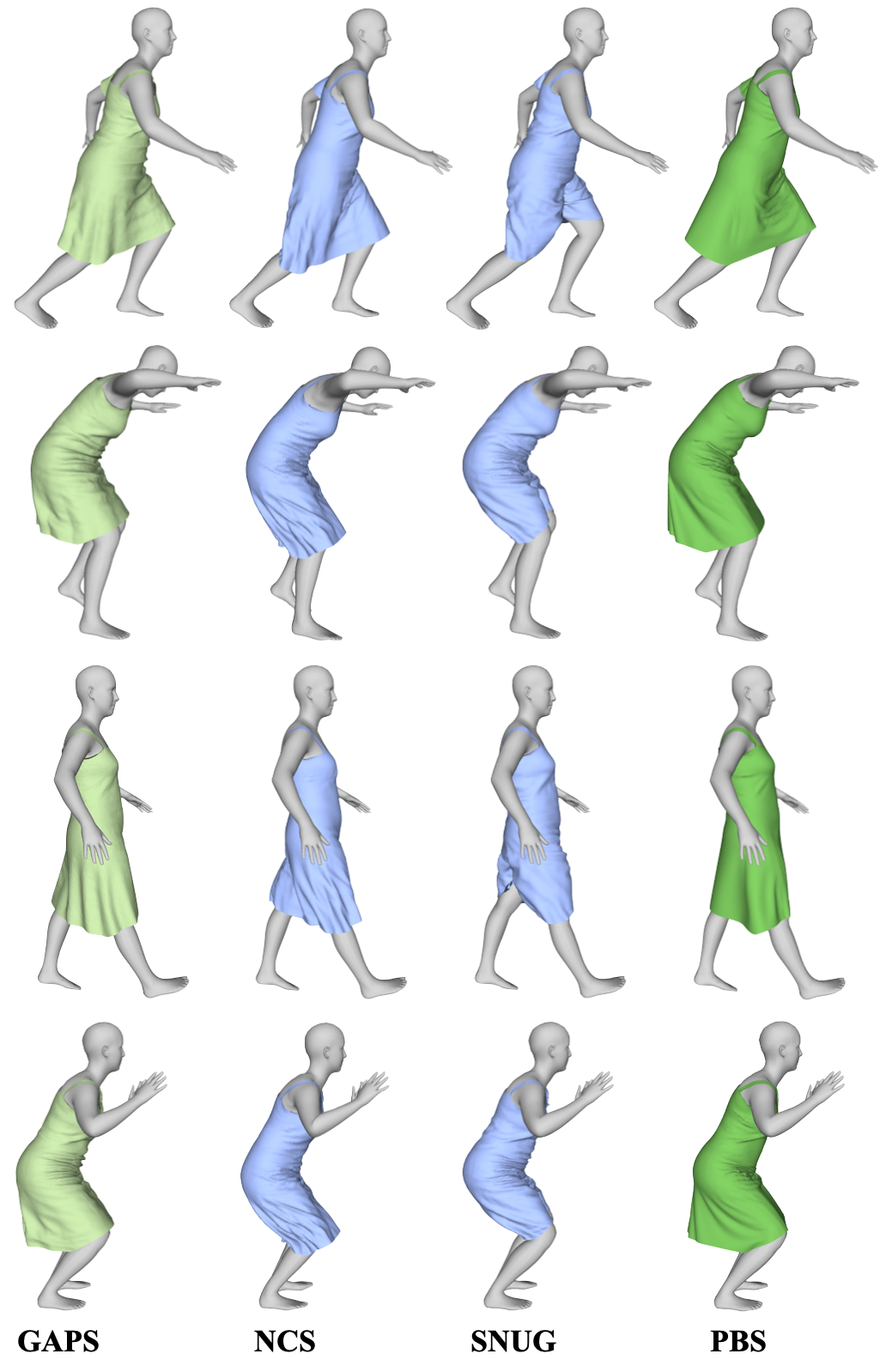}

   \caption{Results on Dress. \textbf{GAPS} shows realistic results quite close to the PBS.}
   \label{fig:skin_weight_quali}
\end{figure}

\begin{figure}[t]
  \centering
   \includegraphics[width=1\linewidth]{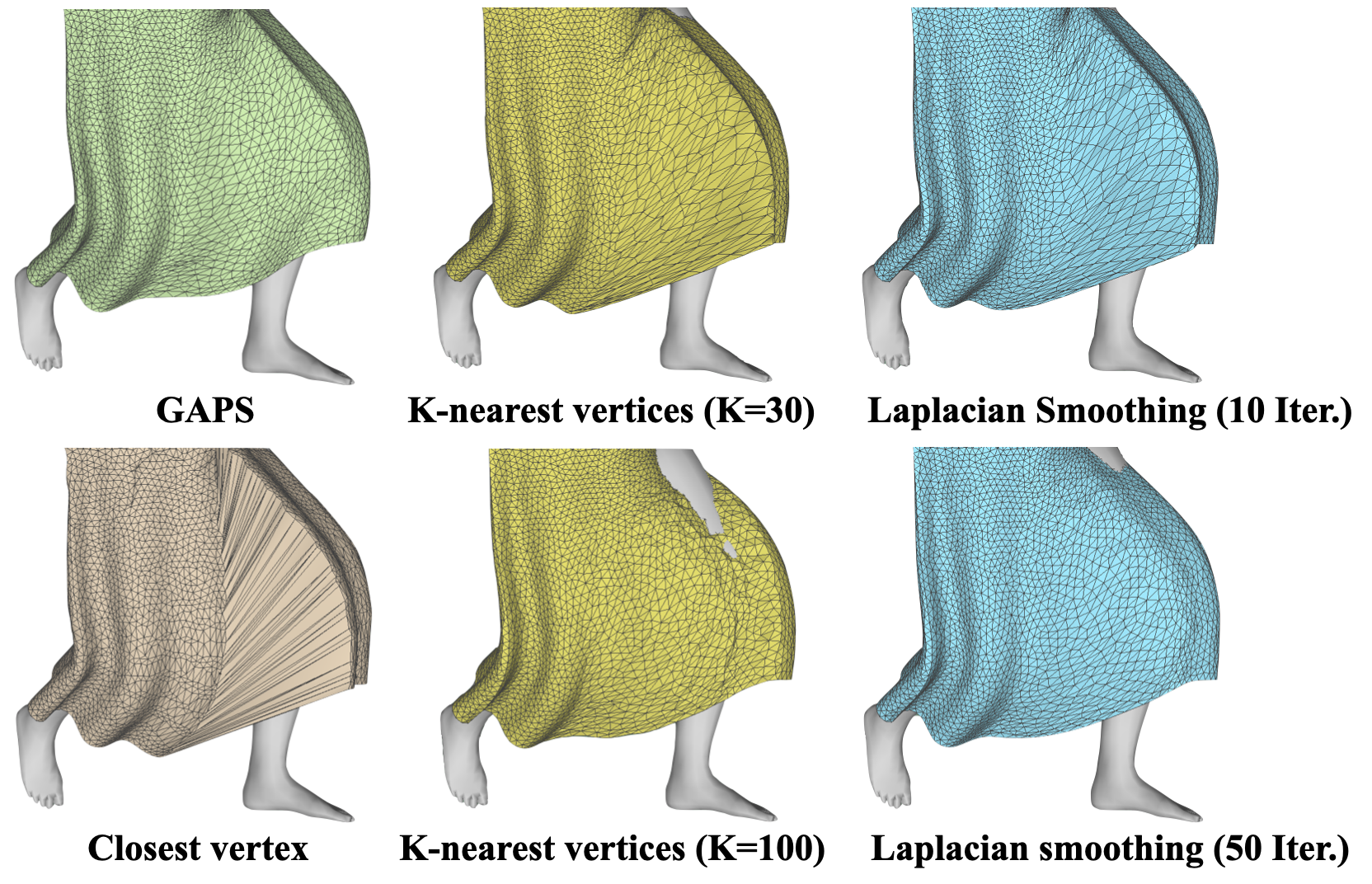}

   \caption{Comparing the geometry-aware, RBF-based skinning in \textbf{GAPS} with others.}
   \label{fig:skin_weight_methods}
   \vspace{-10pt}
\end{figure}

\noindent \textbf{Performance on Loose Garments.}
We revisit dress to compare the performance on loose garments. \textbf{SNUG} assumes that garment motion is close to the body's and uses the blend weights proposed in~\cite{Loper15} for garment vertices and therefore, fails while fitting loose garments. \textbf{NCS} uses a Laplacian smoothing to estimate the blend weights better in order to fit loose garments. However, this requires an iteration parameter to be set. We fine-tune this parameter (set to 50) in order to get the best possible performance on the dress. \cref{fig:skin_weight_quali} shows the results. 
\textbf{SNUG} can not deal with loose garments. On looser areas, we observed a  big spike in strain loss which hinders convergence. \textbf{NCS} can not produce realistic results despite the careful selection of the smoothing iterations; the results are similar to \textbf{SNUG}. In contrast, \textbf{GAPS}' geometry-aware skinning is robust, generates realistic results quite close to the PBS (used in~\cite{Santesteban21}).~\cref{fig:skin_weight_methods} compares skinning of the template garment onto a posed body using different skinning methods. The closest vertex based skinning shows large discontinuities between the legs. The use of K-nearest vertices reduces these discontinuities, however with an inevitable compromise with accuracy due to the fixed number of neighbours. The Laplacian smoothing based skinning of \textbf{NCS} is similar to the k-nearest vertices one, which explains its degraded performance on loose garments.
 
\noindent \textbf{Ablation Study.} We study the individual impact of various components within the \textbf{GAPS} framework.

\textit{1) Collision-awareness.} 
By fixing $k_{\text{ext}}=1$ in $\mathcal{L}_{\text{inext}}$ (see eq.~\eqref{eq:inext_loss}), we impose the garment to remain intextensible everywhere.~\cref{tab:collision} shows the $\%$ of vertices in collision, $\varepsilon_c$. We see that incorporated collision-awareness in
\textbf{GAPS} is crucial for it to drape realistically.

\begin{table}
  \centering
  \resizebox{0.8\columnwidth}{!}{%
  \begin{tabular}{@{}lcccc@{}}
    \toprule
    & T-shirt &  Dress & Skirt & Gown\\
    \midrule
    \textbf{GAPS} & 0.116  & 0.196  &  0.026 & 0.123 \\
    \textbf{GAPS} ($k_{\text{ext}} =1$)  & 5.191  & 5.572 & 2.992 & 7.027 \\
    \textbf{GAPS} without $\mathcal{L_{\text{inext}}}$ & 5.344  & 5.924 & 3.105 & 7.580 \\
    \midrule

  \end{tabular}
   }
  \caption{$\varepsilon_c$ across various configurations shows that collision-awareness is crucial to \textbf{GAPS}. 
  }
  \label{tab:collision}

\end{table}

\textit{2) Inextensibility.}~\cref{tab:isometry} shows that explicit enforcement of inextensibility loss $\mathcal{L}_{\text{inext}}$ on top of physics-based losses $\mathcal{L}_{\text{physics}}$ forces \textbf{GAPS} to maintain inextensibility especially while draping over smaller bodies which are expected to be as inextensible as possible. This shows the complementary nature of strain loss, eq~\eqref{eq:strain} and $\mathcal{L}_{\text{inext}}$~\eqref{eq:inext_loss}.
    
\begin{table}
  \centering
  \resizebox{0.75\columnwidth}{!}{%
  \begin{tabular}{@{}lcccccc@{}}
    \toprule
    & \multicolumn{2}{c}{Avg body} & \multicolumn{2}{c}{Chubby} & \multicolumn{2}{c}{Obese} \\
    & $\varepsilon_{e}$ & $\varepsilon_{a}$ & $\varepsilon_{e}$ & $\varepsilon_{a}$ & $\varepsilon_{e}$ & $\varepsilon_{a}$ \\
    \midrule
    \textbf{GAPS} & 3.05 & 3.83 & 4.32 & 5.06 & 6.41 & 6.44 \\
    No $\mathcal{L_{\text{inext}}}$ & 3.84 & 5.13 & 4.38 & 5.54 & 6.43 & 6.94 \\
    \bottomrule
  \end{tabular}
  }
  \caption{$\varepsilon_{e}$ and $\varepsilon_{a}$ evaluated on Dress across various body types shows the importance of inextensibility loss.}
  \label{tab:isometry} 
  \vspace{-10pt}
\end{table}

\begin{figure}[t]
  \centering
   \includegraphics[width=1\linewidth]{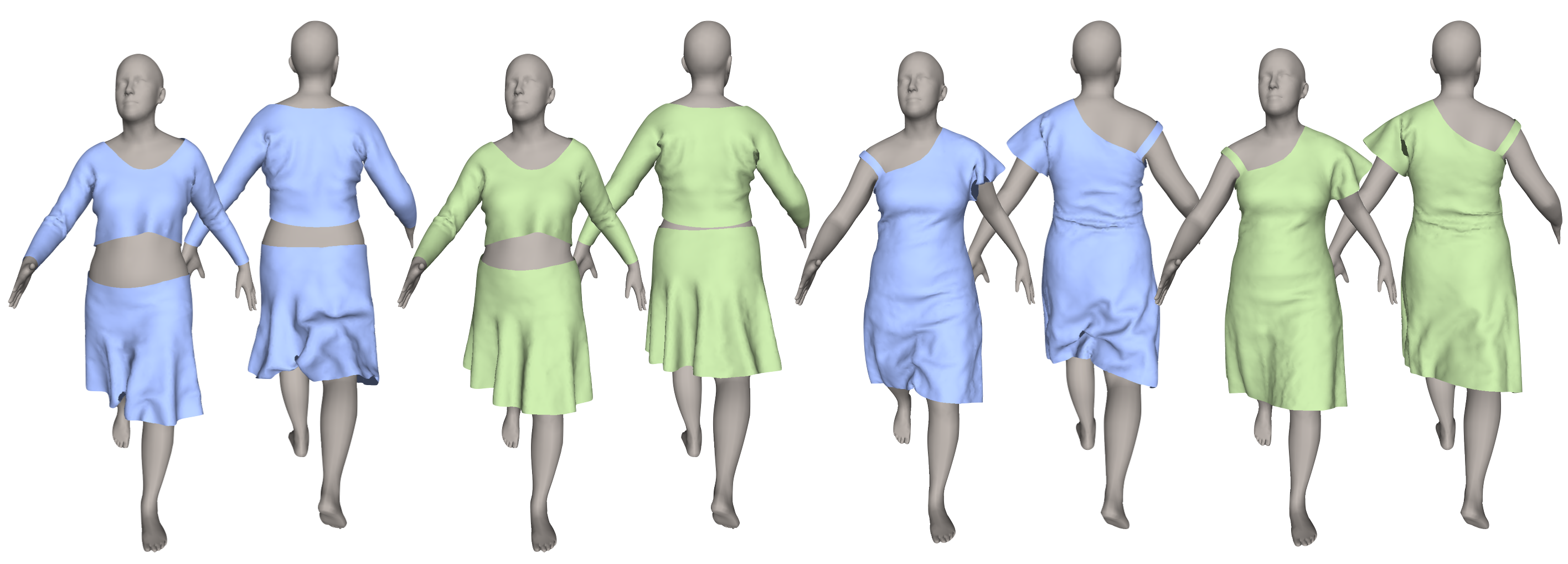}

   \caption{\textbf{GAPS} with (in green) RBF-based skinning and without it (in blue).}
   \label{fig:rbf_skin_abl} 
   \vspace{-5pt}
\end{figure}

\textit{3) RBF-based skinning.} As demonstrated in \cref{fig:rbf_skin_abl}, the RBF-based skinning method employed in \textbf{GAPS} enhances realism for loose garments and generalizes well to tight garments. Note that draping results for tight garments, such as T-shirt, are similar whether the RBF-based skinning method is employed or not.

\noindent \textbf{Timing Performance.}
\textbf{SNUG }takes less than 1 hour to converge for tight garments with less than 10k vertices but will take a significantly longer time (up to 24 hours) for looser garments because of the spike in strain loss.  \textbf{NCS} takes 1-24 hours. Our method takes 1 hour for tight garments and up to 8 hours for looser garments. All training is carried out on 8 NVIDIA V100 GPUs. 

As for run-time performance, we measure the processing duration from the stage of raw body pose data to the final body and garment meshes. For a fair comparison, we use a CMU motion sequence comprising 2,175 frames to evaluate the runtime. The tests are executed on an Intel Xeon Gold 6146 and NVIDIA V100. For \textbf{SNUG}, we used the checkpoint and associated  script provided by the author. For \textbf{NCS}, we use the author's script for training and prediction. \cref{tab:time_perf} shows the comparison.
\textbf{SNUG} takes greater runtime due to the post-processing. The authors suggested the possibility of parallelizing the collision post-processing component on the GPU to enhance runtime performance. However, their runtime GPU code is not publicly available.  
Our approach achieves the fastest performance without requiring a GPU.

\begin{table}
  \centering
  \begin{tabular}{@{}lccc@{}}
    \toprule
    & Train &  Runtime  \\
    \midrule
    \textbf{SNUG}  & 1 - 24 h & 23.7 ms\\
    \textbf{NCS}  & 1 - 24 h & 3.3 ms\\
    \textbf{GAPS} & 1-8 h & 2.7 ms\\
    \bottomrule
  \end{tabular}
  
  \caption{Timing performance. }
  \vspace{-10pt}
  \label{tab:time_perf}
\end{table}

\noindent \textbf{Summary of methods.} \textbf{SNUG} is trained on various body shapes; it is scalable but relies on post-processing to overcome body-garment collisions, which are significant. It cannot handle loose garments. \textbf{NCS} is body-specific and trained on a single body; it is highly restrictive but it fits decently well with lesser collisions and smaller extensions than \textbf{SNUG}. It produces some unrealistic stretches and does not drape loose clothes well, due to its flawed skinning. \textbf{GAPS} is scalable, minimizes body-garment collisions and garment extensions. It produces realistic stretches and drapes all types of garments well.
\section{Conclusion}

We presented \textbf{GAPS}: a geometry-aware, physics-based, self-supervised garment draping method. It incorporates explicit collision-aware inextensibility enforcement which encourages realistic drapings where the garment stretches only while fitting over larger body regions. Furthermore, the collision-awareness module significantly reduces body-garment collisions, eliminating the need for expensive post-processing or restrictive training. In addition, we proposed a geometry-aware skinning approach which automatically computes the right body-participation to a garment dynamics and therefore, can handle a variety of garments. This allows us to obtain significantly better draping results, even on loose garments. Most importantly, our modeling is generic and can be easily incorporated with others. 

\noindent \textbf{Limitations \& future work}
Though temporal neural networks have been incorporated in our approach, there is still a necessity for better handling of the cloth dynamics. Additionally, while the implementation of our collision-aware inextensibility loss has substantially mitigated body-garment collision issues, challenges persist in extreme body poses and shapes. Furthermore, our current framework does not yet address cloth self-collision and cannot deal with multi-layer and topology-varying clothing settings.
As a future work, we would explore the possibility of imposing our collision-aware inextensibility as a hard constraint to ensure zero collisions with minimal garment extensions.

\noindent \textbf{Acknowledgements} This research has received funding from the project RHINO, an ANR-JCJC research grant.

{
    \small
    \bibliographystyle{ieeenat_fullname}
    \bibliography{main}
}
\end{document}